\documentclass{svproc}
\usepackage{url}

\usepackage{amsmath}
\usepackage{amssymb}
\usepackage{graphicx}
\usepackage{algorithm}
\usepackage{algorithmic}
\usepackage{booktabs}
\usepackage{wrapfig}
\usepackage{multicol}
\usepackage{multirow}
\usepackage{hyperref}

\begin{document}
\mainmatter              
\title{Privileged Reinforcement and Communication Learning for Distributed, Bandwidth-limited Multi-robot Exploration}
\titlerunning{Privileged RL and CL for Bandwidth-limited Multi-Robot Exploration}  
%
\author{Yixiao Ma\inst{1,2} \and Jingsong Liang\inst{1,2} \and
Yuhong Cao\inst{2} \and Derek Ming Siang Tan\inst{2} \and Guillaume Sartoretti\inst{2}}
\authorrunning{Yixiao Ma et al.} 
%
\tocauthor{Yixiao Ma, Jingsong Liang, Yuhong Cao, Derek Tan,
and Guillaume Sartoretti}
\institute{School of Computing, National University of Singapore, SG 117417.\\
\email{yixiaoma@u.nus.edu, jingsongliang@u.nus.edu},
\and
Mechanical Engineering Dept., National University of Singapore, SG 117575.\\
\email{caoyuhong@nus.edu.sg, derektan@u.nus.edu, mpegas@nus.edu.sg},\\
WWW home page: \texttt{http://www.marmotlab.org}}

\maketitle              

\begin{abstract}

Communication bandwidth is an important consideration in multi-robot exploration, where information exchange among robots is critical. While existing methods typically aim to reduce communication throughput, they either require significant computation or significantly compromise exploration efficiency. In this work, we propose a deep reinforcement learning framework based on communication and privileged reinforcement learning to achieve a significant reduction in bandwidth consumption, while minimally sacrificing exploration efficiency. Specifically, our approach allows robots to learn to embed the most salient information from their individual belief (partial map) over the environment into fixed-sized messages. Robots then reason about their own belief as well as received messages to distributedly explore the environment while avoiding redundant work. In doing so, we employ privileged learning and learned attention mechanisms to endow the critic (i.e., teacher) network with ground truth map knowledge to effectively guide the policy (i.e., student) network during training. Compared to relevant baselines, our model allows the team to reduce communication by up to two orders of magnitude, while only sacrificing a marginal 2.4\% in total travel distance, paving the way for efficient, distributed multi-robot exploration in bandwidth-limited scenarios. We open-sourced our full code~\footnote{https://github.com/marmotlab/Bandwidth-Limited-Multi-Robot-Exploration}..

\keywords{Deep Reinforcement Learning, Communication Learning, Multi-robot exploration, Distributed Path Planning}
\end{abstract}


\section{Introduction}

Information sharing in multi-robot exploration is critical to generating high-quality, distributed exploration paths~\cite{wang2022distributed}: robots must communicate with each other to obtain more information beyond their own partial knowledge, to make cooperative decisions that can speed up task completion and avoid redundant work.
Under normal circumstances, it is usually viable for most exploration planners to share vast amounts of important information among team members (such as full partial maps, or robot trajectories) continuously during the mission.
However, these approaches become impractical in bandwidth-constrained settings such as underwater and underground environments~\cite{kida2018experimental,salam2020underground}, where communication range is often inversely linked to communication bandwidth.
Recent research shows that the transmission of an occupancy grid map usually requires a bandwidth of $\sim 2$ Mbps~\cite{tabib2019real}, but that underwater communication bandwidth rarely exceeds 100kbps~\cite{kida2018experimental}.
This problem of bandwidth constraints is further exacerbated in larger teams, where (even modern) communication channels may not be able to withstand high communication throughput between numerous robots~\cite{klavins2004communication}.

\begin{figure}[t]
\vspace{-5pt}
\centering
\includegraphics[width=0.8\linewidth]{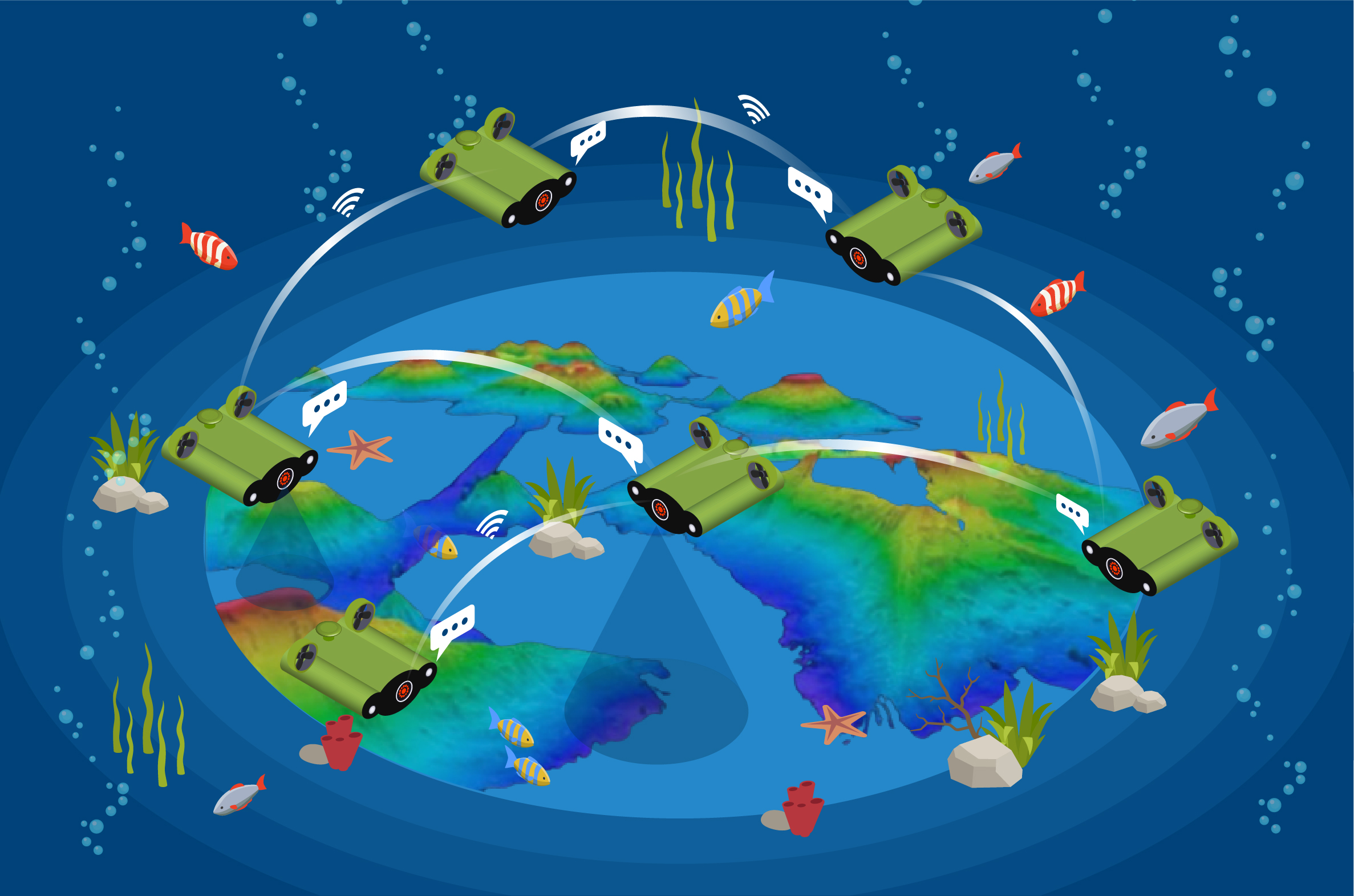}
\vspace{-0.4cm}
\caption{Example application of our approach to a multi-robot exploration task in a bandwidth-constrained environment (here, underwater).}
\label{fig: intro}
\vspace{-10pt}
\end{figure}

There are currently two main strategies to reduce communication throughput among robots. The first strategy is to simply decrease the frequency of communications~\cite{masaba2021gvgexp}. However, this strategy often leads to poor cooperation, mainly due to the lack of up-to-date information from other robots. The second strategy is to reduce the size of messages, e.g., by only sharing low-dimensional representations of each others' partial map, which recipients can then process to reconstruct/estimate the full map~\cite{javanmardi2017autonomous,zhang2022mr}. However, these approaches usually come at important computational costs and may result in lower exploration efficiency when essential details from the original map are lost during map exchange.

To address these problems, we propose a novel DRL-based multi-robot exploration framework based on communication learning and privilege learning, tailored for bandwidth-limited scenarios.
Our framework relies on learned messages as an alternative to conventional map sharing.
Our approach primarily relies on communication learning to allow robots to learn to encode their own belief map into a small, fixed-sized message that is shared with other robots.
In doing so, our communication layer allows robots to learn to identify, encode, and share the most salient portion of their individual belief with each other, within given constraints over message length (i.e., maximum bandwidth within the system).
Robots then learn to reason about their own knowledge/state as well as receive messages to form an implicit representation of the overall explored environment.
This enables the generation of high-quality, distributed exploration paths.
Following our recent work in single robot exploration~\cite{cao2024deep}, we rely on privileged learning to boost the performance of our final model.
Specifically, we let our critic network access ground truth information during training only, allowing it to provide more accurate action evaluation for the training of the robots' policy network.
This training approach significantly enhances our final model's long-term planning capabilities, by allowing robots to reason about their knowledge, as well as received messages, at different spatial and temporal scales.

We compare our model to a conventional multi-robot exploration planner in a set of 100m$\times$100m indoor maps and investigate the impact of using learned messages over traditional partial map sharing.
Our results show that our model can reduce the volume of communications by up to \textbf{99.2\%}, at the cost of a marginal \textbf{2.4\%} performance loss.
These results highlight the capability of our robots to understand and reason about the current global state of the exploration task, without explicitly relying on other robots' detailed map.
We finally train a variant of our DRL-based model, where robots are allowed to both share learned messages and explicit partial maps.
This model outperforms our map-sharing-free approach and the conventional baseline by 11.4\% and 9.2\% respectively in terms of exploration distance, highlighting the power of our general framework in high-bandwidth scenarios where full maps may be reliably shared among robots.


\section{Prior Works}


\subsection{Multi-robot Exploration}

Approaches to conventional multi-robot exploration are methodologically classified into two main categories: frontier-based and sampling-based. For example, Yu et al.~\cite{yu2021smmr} applied artificial potential fields to attract robots towards diverse frontiers while ensuring mutual repulsion to maintain distance between each other.
Most recently, Cao et al.~\cite{cao2023representation} proposed an mTSP (Multiple Traveling Salesman Problem) based global planner in conjunction with a sampling-based local planner to explore large-scale environments.
The centralized global planner segments exploration areas into several vital nodes and then assigns them to robots.
However, these approaches remain greedy and prioritize short-term efficiency, often resulting in shortsighted path planning and suboptimal performances.

Given the rapid advancement of neural networks, many works have looked to deep learning to enhance autonomous exploration.
Niroui et al.~\cite{niroui2019deep} first pioneered the integration of frontier-based methods with deep reinforcement learning to resolve short-sightedness problems in single-robot exploration.
Yu et al.~\cite{yu2023asynchronous} employed asynchronous multi-robot proximal policy optimization as a training method to address multi-robot exploration.
Concurrently, Luo et al.~\cite{luo2019multi} leveraged graph convolutional neural networks to enhance the efficiency of multi-robot exploration.
Our most recent work proposed ARiADNE~\cite{cao2023ariadne,cao2024deep}, an attention-based neural network to achieve state-of-the-art in long-term planning for single-robot exploration.
In this work, we extend the ARiADNE framework to multi-robot exploration under bandwidth limitations.


\subsection{Communication in Multi-Robot Exploration}

Many existing methods~\cite{luo2019multi,yu2023asynchronous,yu2021smmr} adopt continuous communication to gather up-to-date information that assists robots in making cooperative decisions.
However, this often comes at the cost of high communication bandwidth requirements.
Some methods reduce communication throughput by decreasing the frequency of communications.
For example, Masaba et al~\cite{masaba2021gvgexp} proposed a method to utilize a generalized Voronoi graph to segment the environment into sub-regions and allocate robots into distinct regions to explore.
Robots only communicate with each others when their own region of responsibility has been fully explored, at which point a new region will be allocated to them.
However, this method represents a trade-off between exploration efficiency and communication frequency, because the lack of real-time information from other robots often results in poor cooperation. Alternatively, some methods chose to reduce the size of messages. For instance, Gaussian Mixture Model (GMM) Maps~\cite{javanmardi2017autonomous} involves exchanging key map parameters and then attempting to reconstruct the map.
Nevertheless, these approaches require substantial computational resources, hindering their practical implementation for exploration tasks that demand real-time decision-making.
Another example~\cite{rekleitis1999efficient,zhang2022mr} extracts semantic information from the map to create a compact topological map that is shared with other robots.
However, this approach often results in lower exploration efficiency due to the loss of some details from the original map in the topological representation.


\subsection{Communication Learning}

Foerster et al.~\cite{foerster2016learning} introduced the concept of communication learning in multi-robot reinforcement learning: robots learn what information/message to share within the team, thereby effectively augmenting the knowledge available to individual robots and mitigating the challenges posed by partial observability.
Furthermore, Freed et al.~\cite{freed2020communication} introduced a differentiable discrete binary channel for communication learning and demonstrated that this approach surpasses the robustness and performance of the reinforced communication learning~\cite{foerster2016learning} proposed by Foerster et al. 
The key component in communication learning is message generation. Messages can be generated through two methods: by inputting each robot's observation into a message generation module~\cite{foerster2016learning,kim2019learning,kim2020communication} or by sequentially processing a message that is circulated among the robots~\cite{peng2017multiagent,wang2023full}.
Specifically, Kim et al.~\cite{kim2020communication} introduced a novel scheme of intention-sharing within communication learning by swapping intentions that are encoded from partial observations.
On the other hand, FCMNet\cite{wang2023full} and FCMTran~\cite{wang2023full} let robots communicate with each other by enabling sequential message processing among robots, achieving state-of-the-art status in StarCraft II Micromanagement benchmarks.
These works demonstrate that communication learning can enable robots to reconstruct and utilize information from high-dimensional compressed messages.


\section{Problem Formulation}

In this work, we consider multi-robot exploration under constraints over communication bandwidth.
We let $W$ be a bounded environment, which can be divided into free area ${W}_{f}$ and occupied area $W_{o}$, where $W_{f}\cup{W}_{o}=W$ and $W_{f}\cap{W}_{o}=\varnothing$.
The belief map $M_{i}$ is locally maintained by the $i$-th robot about ${W}$, and is categorized into three components: free area $M_{i}^{f}$, occupied area $M_{i}^{o}$, and unknown area $M_i^u$.
Initially, each robot starts at $p_i^0$ with an empty individual belief map, such that ${M_i}={M_i^u}$. During the exploration task, each robot makes individual decisions and generates/follows a trajectory, i.e., a sequence of waypoints $\psi_i = \{p_i^0,p_i^1,...\}$, each located in the free area ${W}_{f}$.
Each robot updates its belief map according to regular measurements along its path, by (re-)classifying formerly-unknown area into either ${M}_{f}$ or $M_{o}$.

The objective of multi-robot exploration can be formulated as a MinMax function, where we aim to minimize the maximum trajectory length among all robots (i.e., makespan of the task):
\begin{equation}
    \label{eq:objective}
    \min \max_{i\in{1,...,n}} L(\psi_i), 
\end{equation}
where $L(\psi_i) = \sum_{t=1}^{m}D(p_i^{t-1}, p_i^{t})$ is the trajectory length to complete exploration, with $D(p_i^{t-1},p_i^{t})$ the distance between $p_i^{t-1}$ and $p_i^{t}$.
We consider the multi-robot exploration completes when the exploration rate (i.e., the proportion of explored free area $\lvert\bigcup_{i=1}^{n}M_i^f\rvert$ compared to the true free area in the environment $W_f$) exceeds a given threshold $\theta$.

In this work, we further consider a bandwidth constraint for robots' communication during exploration, preventing robots from sharing their local belief map $M_i^t$ with each others (in practice, we impose a maximum of \~256 bytes per message).
Instead, robots can only exchanges compact messages, containing small amounts of information such as their current position.
Note that, without accurate map sharing, robots cannot fully confirm whether an area in the environment has been explored by others or not.
Therefore, in such a problem setup, it is more reasonable to set the exploration objective as efficiently exploring most of the environment rather than the whole environment.
Consequently, we set the exploration terminating threshold $\theta$ between $[0.9, 1]$ in practice.


\section{Distributed Bandwidth-Limited Exploration}

\begin{figure}[t]
  \vspace{-5pt}
  \centering
  \includegraphics[width=1\linewidth]{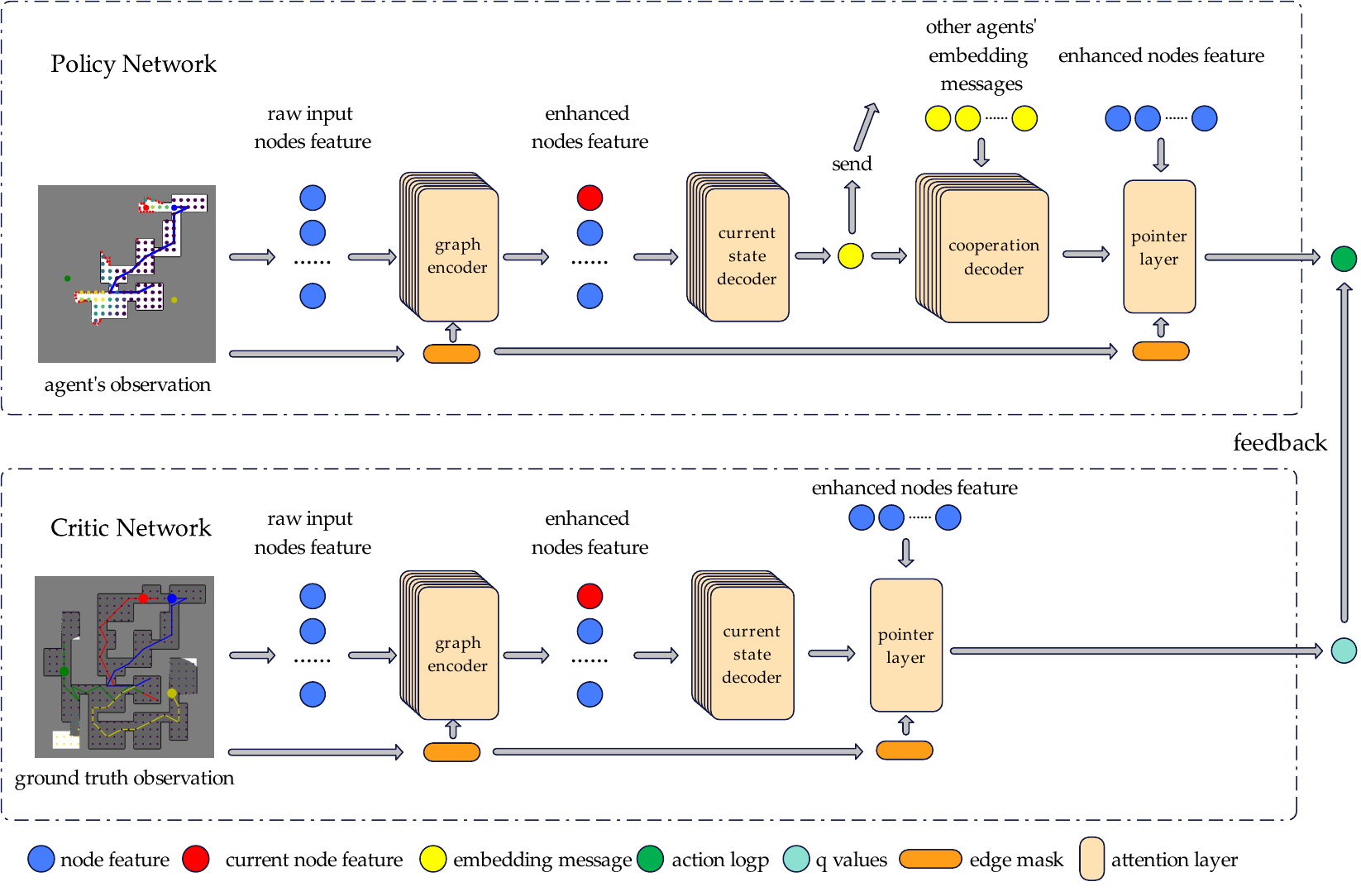}
  \vspace{-0.8cm}
  \caption{\textbf{Architecture of our policy and critic networks.}}
  \label{fig: network}
  \vspace{-0.3cm}
\end{figure}


\subsection{Exploration As An RL Problem}

We cast the exploration problem as a partially observable Markov Decision Process, formulated using the tuple $(S,A,\Gamma, R,\Omega, O)$, where $S$ denotes the state space, $A$ the action space, $\Gamma$ the state transition function, $R$ the reward set, $\Omega$ the observation space of the robot, and $O$ the observation function. During exploration, the robot does not have access to the true state $s^t, \forall{s^t}\in{S}$; instead, it operates based on its observation $o^t, \forall{o^t}\in{\Omega}$. According to its observation, each robot selects and executes action $a^t\sim\pi(\cdot \mid o^t)$, where it selects and travels to the next way-point $p^{t}$. The next state $s^{t+1}$ is determined by the current state $s^t$ and the action $a^t$, through the state transition function $\Gamma(s^{t+1}\mid{s^t},a^t)$ (unknown to the ). 
Each robot$_i$'s objective is to learn an optimal policy $\pi^*(a_i^t \mid o_i^t)$ that maximizes the long-term expected return $\mathbb{E} \left [ \sum_{t=1}^{m}r^t \right ] $, where $r^t\in{R}$.
In the following sections, we detail the observation, action space, and reward structure of our RL formulation. 


\subsection{Policy Network}

\subsubsection{Policy Network Observation}

Similar to our prior works~\cite{cao2023ariadne,cao2024deep}, the observation for each robot is an informative graph $G^{t}=(V^{t}, E_i^t)$ defined over the free explored space in each robot's partial map. $E_i^t$ is defined as the collision-free edges connecting each vertex in  $V^{t}$ with its k nearest neighbors. Each vertex $v_i=(c^t_x, c^t_y, u^t, g^t, p^t) \in V^{t}$ has four components, namely coordinates, utility, guidepost, and position indicator, where coordinates $(c^t_x, c^t_y)$ represent the 2D position of the graph vertex, utility $u^t$ measures the number of observable frontiers from that vertex, and the guidepost $g^t$ component represents whether this vertex has been visited previously. 
Extending~\cite{cao2023ariadne} to a multi-robot setting, we further augment the observation with an additional component called occupancy, which indicates whether each vertex is currently occupied by the current robot, by another robot, or is unoccupied, with values -1, +1, and 0 respectively.  
This component enables each robot to understand the geometric relationships between their position and those of other robots, thereby improving their cooperative capabilities.
To represent another robot located outside the area explored by the robot so far (i.e., outside the area covered by its graph $G^{t}$), we manually insert a temporary vertex $v$ to the graph $G^{t}$.

\subsubsection{Graph Encoder}

For each robot, we use a graph encoder to extract high-dimensional and informative node features from its informative graph. The graph encoder contains multiple consecutive attention layers~\cite{vaswani2017attention}, each learning to intelligently fuse neighboring node features.
We first project every node from $V^{t}$ in the informative graph $G^{t}$ into a high-dimensional (dimension $d=64$ in practice) node feature. 
We denote those raw node features as $F^{raw} \in \mathbb{R}^{d\times{N}}$, where $N$ is the number of nodes in the graph.  
Thereafter, we input $F^{raw}$ and the edge mask $\hat{M}$ into attention layers. The edge mask $\hat{M} \in \mathbb{R}^{N \times N}$ is calculated from $E_i^t$, where 
$\hat{M}_{ij}= \left \{ \begin{array}{cc}
     & 0, (v_i,v_j) \in E_i^t \\
     & 1, (v_i,v_j) \notin E_i^t.
\end{array} \right .$. It is applied in the attention layers to only allow the merging of neighboring node features. 
Each attention layer reads:
\begin{equation}
    \begin{array}{cc}
        & q_i = W^Q h_i^q, k_i = W^K h_i^{k,v}, \\
        & v_i = W^V h_i^{k,v}, u_{ij} = \frac{q_i^T \cdot k_j}{\sqrt{d}}, \\
        & w_{ij} = \left \{ \begin{array}{cc}
             &  \frac{e^{u_{ij}}}{\sum_{j=1}^n e^{u_{ij}}}, \hat{M}_{ij}=0 \\
             &  0, \hat{M}_{ij}=1
        \end{array} \right . ,
        h'_i=\sum_{j=1}^n w_{ij} v_j,
    \end{array}
    \label{eq:attention}
\end{equation}
where $W^Q, W^K, W^V \in \mathbb{R}^{d \times d}$ denote learnable weight matrices, $h^q$ is the query source, and $h^{k,v}$ is the key-and-value source. Note that in the graph encoder, $h^q=h^{k,v}$ (known as self-attention). For the first attention layer, $h^q=h^{k,v}=F^{raw}$. The following layers sequentially take the previous layer's output as input. We obtain the output of the final attention layer as enhanced node features $\hat{F} \in \mathbb{R}^{d \times N}$.

\subsubsection{Current State Decoder}

The current state decoder distills the relationships between the current robot position and all other graph vertices.
We first input the enhanced node features of the current node $\hat{F}_c$ as the query source and all enhanced node features $\hat{F}$ as the key and value source into an attention layer. 
Thereafter, we concatenate the output features together with $\hat{F}_c$ and re-project the output to a d-dimensional feature $\mu$. 
Here, $\mu$ represents the knowledge about current accessible state (not global state, because our robot's observation is partial), capturing multi-scale dependencies among previously traversed areas.
Finally, $\mu$ is broadcast to all other robots as this robot's message for this decision step, as we discussed in section 3.

\subsubsection{Cooperation Decoder}

The cooperation decoder aggregates relationships between an agent's own learned message (output of the current state decoder) and the received learned messages from other robots to form an implicit representation of the overall explored environment.
We first take robot$_i$'s learned message $\mu_i^t$ as the query source of a cross-attention layer. Concurrently, we concatenate all other robots' learned messages, $\mu_{-i}^t = \mu_{1:n}^t \setminus \mu_i^t$, and pass it into the same attention layer as the key-and-value source.
From here, we derive the cooperative feature $F^{co} \in \mathbb{R}^{d \times 1}$ , which represents the implicit context of the overall explored environment reasoned from all robots' learned messages.
Notably, the cooperation decoder can accept messages from an arbitrary number of robots, highlighting the scalability of our approach to different team sizes.

\subsubsection{Action Space}

The action space comprises all neighboring vertices to the current vertex the robot is occupying.
We pass the cooperative feature $F^{co}$ and the neighboring nodes enhanced feature $\hat{F}_{nbr}$ into the pointer layer~\cite{cao2023ariadne,cao2022dan,liang2023context}.
Its output takes the form of a logarithmic probability distribution, from which we sample an action during training, and select an action greedily during testing/deployment.
Overall, the attention-based neural network outputs a stochastic policy described by $\pi(a_i^t \mid o_i^t) = p_i^{t+1}, (p_i^{t+1}, p_i^t) \in E_i^t$.


\subsection{Training}

We employ parameter sharing to train a homogeneous policy for all agents.
Specifically, we rely on the soft actor-critic (SAC) algorithm~\cite{christodoulou2019soft,haarnoja2018soft}, which trains a policy networks and double critic networks.
Compared to standard SAC, following our previous work~\cite{cao2024deep}, we further propose a privileged critic network which significantly boosts the performance of the trained model.

\subsubsection{Privileged critic network}

\begin{figure}[t]
  \centering
  \includegraphics[width=1\linewidth]{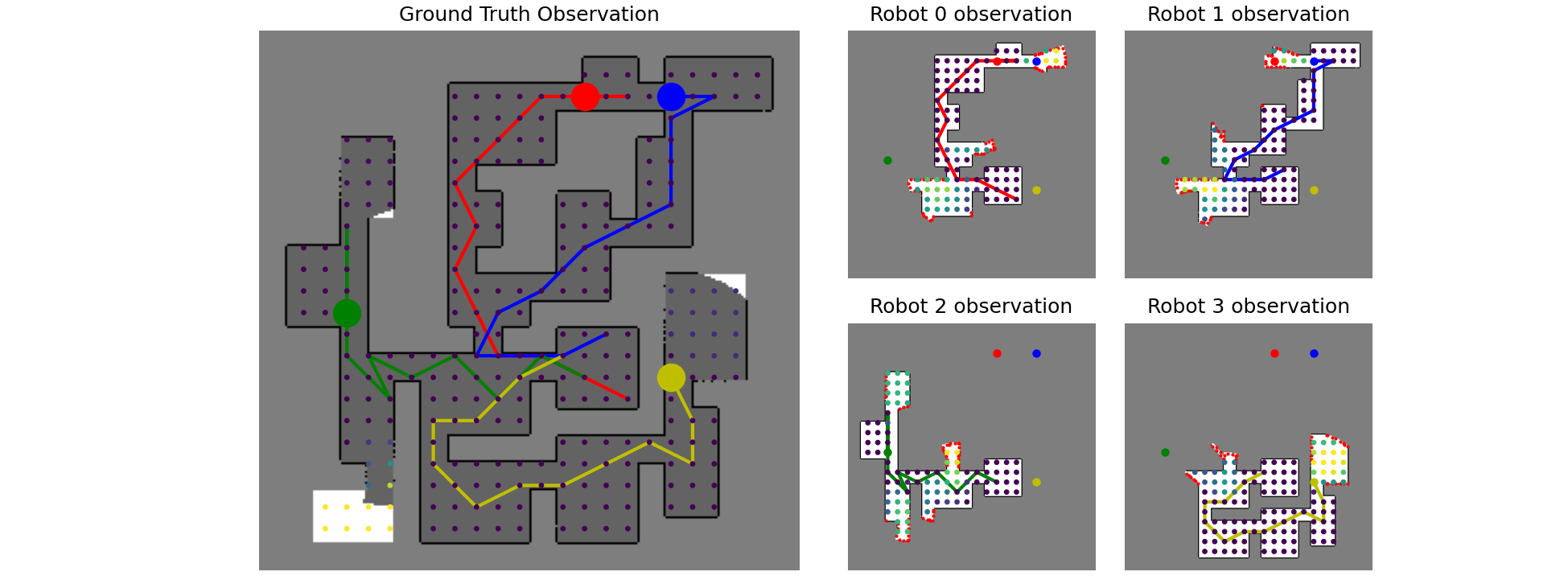}
  \caption{\textbf{Observation structure for our policy and critic networks. } \scriptsize{The left part is the ground truth observation used in our critic network, while the four plots on the right respectively represent the partial observations used as input to the robots' policy networks. The large dots in red, blue, yellow, and green represent the current position of the robots, while the smaller nodes' varying colors denote varying exploration utility. We omit the depiction of the collision-free graph edges for visualization purposes.}}
  \label{fig: Observation}
\end{figure}

We employ privileged learning to endow the critic (i.e. teacher) model with ground truth map knowledge to effectively guide the policy (i.e. student) model during training. 
It aims to increase the accuracy of the state-action values (the predicted long-term return) predicted by the critic network, which is then used to guide the policy network training. 
In particular, compared to our policy network discussed in the previous section, we modify the observation and network structure of the critic network. 
First, our critic network utilizes a ground truth graph $\hat{G}^t=(\hat{V}^{t}, \hat{E}^t)$ (derived from the ground truth state $s^t$) as the observation instead of the partial informative graph $G^{t}$  (derived from belief map $M_i^t$), as illustrated in Figure~\ref{fig: Observation}. 
There, utilities of nodes in $\hat{V}^{t}$ are determined by the number of neighboring unexplored free area.
\begin{wrapfigure}{r}{0.4\textwidth}
\vspace{-0.4cm}
  \centering
  \includegraphics[width=1\linewidth]{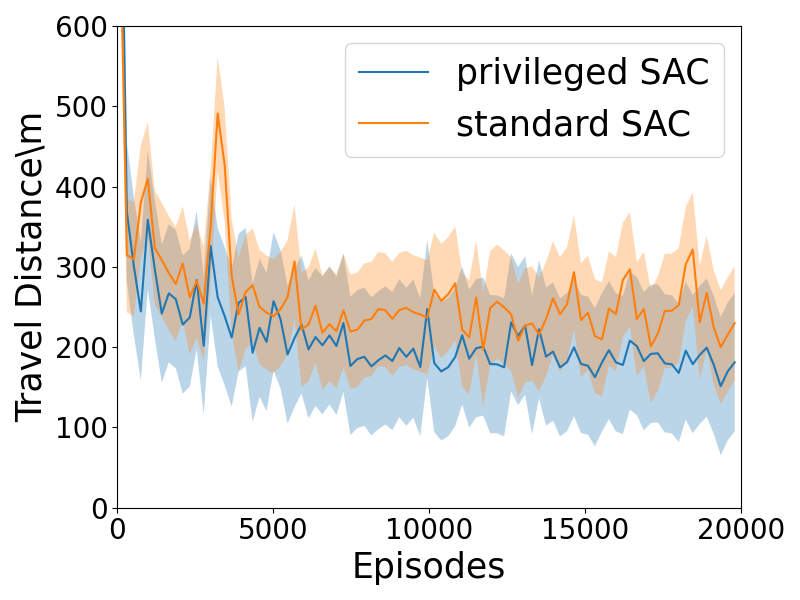}
  \caption{\textbf{The ablation of Privileged Learning. } \scriptsize{Both models have been trained for 20,000 episodes to converge. A lower travel distance signifies better model performance.}}
  \label{fig: ablation_sac}
  \vspace{-1cm}
\end{wrapfigure}
Second, we eliminate the communication layer from the critic network. As the model has access to the global ground truth already, messages shared by other agents are not helpful for return estimation.  
As seen from figure~\ref{fig: ablation_sac}, the introduction of privileged map information significantly enhanced training performance by 23.8\%.
Moreover, we notice improved training stability as seen from lower variance compared to the standard SAC.
This demonstrates that the critic network can better guide the policy network with more accurate long-term return estimation derived from privileged information.

\subsubsection{Reward Structure}
We modify our training reward structure based on our previous single robot exploration work~\cite{cao2023ariadne} to encourage robots to collaboratively explore the environment. The original reward structure in~\cite{cao2023ariadne} consists of three parts: (1) The first part $r_o$, is the is the number of observed frontiers at the new viewpoint. (2) The second part $r_c$ is a punishment on the distance between the previous and new viewpoints. (3) The last part $r_f$ is a finishing reward given after the environment has been fully explored.
In our approach, we perform three modifications to this original reward function.  
First, we change $r_o$ to the unexplored area around the new viewpoint using the ground truth map. This aims to provide a more accurate assessment of the exploration utility given the selected position. 
Second, we introduce a team reward $\xi$, which is related to the newly explored area discovered between the previous and current viewpoints (i.e., $\xi^t = \rho (\bigcup_{i=1}^{n}M_i^{t+1} \setminus \bigcup_{i=1}^{n}M_i^t)$, where $\rho$ is a normalize factor). This reward is designed to encourage robots to explore cooperatively instead of greedily pursuing individual exploration reward $r_o$ .  
Third, we introduce a momentum reward $\lambda_i^t$ to encourage robots to move in any general direction with minimal backtracking. Its purpose is to shorten the highly random processes at the beginning of training and optimize robot's trajectory. The calculation of $\lambda_i^t$ is detailed in algorithm~\ref{alg:momentum}. Thus, the overall reward at decision step $t$ is 
\begin{equation}
    r^t(a^t, o^t) = r_o + \lambda_i^t + \xi^t + r_f - r_c,
    \label{eq:reward}
\end{equation}

\begin{algorithm}
    \caption{momentum reward calculation}
    \label{alg:momentum}
    \begin{algorithmic}[1]
        \REQUIRE  current position $p^t = (x_1,y_1)$, next way-point $p^{t+1} = (x_2, y_2)$, previous step movement direction $\Vec{\alpha}$;\\
    
        \STATE current step movement direction $\Vec{\beta} \leftarrow p^t \cdot p^{t+1} = (x_1 x_2, y_1 y_2)$; \\
        \STATE momentum reward $\lambda^t \leftarrow 0.1 \times (\Vec{\alpha} \cdot \Vec{\beta})$; \\
        \STATE $\Vec{\alpha} \leftarrow \Vec{\beta}$; \\
    
        \STATE \textbf{Return:} $\lambda^t$
    \end{algorithmic}

\end{algorithm}


\subsubsection{Training Details}

Our model is trained on a set of 4000 dungeon maps like figure~\ref{fig: Observation}, each $100m \times 100m$. The sensor range of our robots is 20 meters. Nodes are uniformly distributed 4 meters apart in the free area. Each node connect to its neighboring nodes up to 12 meters away (i.e., max 24 neighboring nodes). During training, we set the number of robots $N=4$, the max episode length is set to 128 steps, the batch size to 256, the learning rate to $1e-5$, and discount factor $\gamma=1$. We set the episode buffer size to 20000, and commence training when at least 5000-step data have been collected. We utilize Ray~\cite{moritz2018ray} for parallel data collection. The model requires around 20,000 episodes to converge, taking around 108 hours on a desktop with an Intel i9-9900k CPU and two NVIDIA RTX 2080 SUPER GPUs.


\section{Experiments}


\subsection{Experimental Setup}

We extend TARE~\cite{cao2021tare}'s TSP-based local planner to multi-robot settings, and refer to it as \textbf{mtsp\_based}.
Specifically, the resulting \textbf{mtsp\_based} planner samples the frontiers into viewpoints and connects them through solving a mTSP (in practice we use ORtools).
Note that this exploration planner is distinct from the multi-robot planner used in MUI-TARE~\cite{yan2023mui} and MTARE~\cite{cao2023representation}: MUI-TARE~\cite{yan2023mui} specifically aims at enhancing the quality of map merging during the exploration process, while MTARE~\cite{cao2023representation} focuses on exploration under communication constraints in large-scale environments. Therefore, we adopted $mtsp\_based$ as our baseline, as a better alternative to fairly represent conventional algorithms.

In \textbf{mtsp\_based}, each robot shares its own belief map with other robots, forming a global belief map with positions  $(\bigcup_{i=1}^{n}M_i^t,p_{1:n}^t)$.
On the other hand, for a fair comparison of the long-term exploration capabilities, we proposed a variant of our model: $global\_map$, which also has access to shared belief maps.
That is, its observation is $(\bigcup_{i=1}^{n}M_i^t,\psi_i^{1:t},p_{1:n}^t,\mu_{1:n}^t)$ instead of the original observation $(M_i^t,\psi_i^{1:t},p_{1:n}^t,\mu_{1:n}^t)$.

In our experiments, robots communicate through learned messages at every time step $t$, where each message packet is a 1x64-dimensional float32 tensor (i.e., $64 \times 4$ bytes). Our competing baselines communicate using belief map messages, where each message packet is a $H \times W$ int8 tensor (i.e., $H^t \times W^t \times 1$ bytes), and $H$ and $W$ is the height and width of the map message respectively.

\begin{table}[t]
   \tiny
    \caption{\textbf{Comparisons with baselines ($\theta=0.95$).}\small{The criteria for evaluation includes the average and standard deviation of travel distance (abbreviated as D), upload data volume (abbreviated as UV), download data volume (abbreviated as DV) and total planning time (abbreviated as T). Lower values for these metrics are preferable ($\downarrow$). When collecting statistics, we report the performance of the worst-performing robot within each scenario.}}
    \label{tab:exp_0.95}
    \centering
        \begin{tabular}{cccccccc}
        \toprule
        \multirow{2}{*}{\scriptsize{Nums of robot}}& \multicolumn{2}{c}{\scriptsize{mtsp\_based}} & \multicolumn{2}{c}{\scriptsize{global\_map}} & \multicolumn{2}{c}{\scriptsize{ours}}\\
        \scriptsize{} & \scriptsize{4} & \scriptsize{8} & \scriptsize{4} & \scriptsize{8} & \scriptsize{4} & \scriptsize{8}\\
        \midrule
        \scriptsize{D}($m$)$\downarrow$  
             & 200.8($\pm 44.0$) & 182.2($\pm 41.7$)  
             & \textbf{182.2}($\pm 35.6$) & \textbf{150.1}($\pm 47.5$)  
             & 205.7($\pm 64.6$) & 168.9($\pm 84.45$) \\
             
        \scriptsize{UV}(MB)$\downarrow$  
             & 656.2($\pm 206.8$) & 554.7($\pm 191.4$)  
             & 679.2($\pm 174.1$) & 516.6($\pm 230.9$)   
             & \textbf{5.7($\pm 1.8$)} & \textbf{4.5($\pm 2.2$)} \\
             
        \scriptsize{DV}(MB)$\downarrow$         
             & 1791.1($\pm 566.0$) & 3303.1($\pm 1146.9$) 
             & 2037.7($\pm 522.5$) & 3616.4($\pm 1616.8$)
             & \textbf{17.2($\pm 5.5$)} & \textbf{32.0($\pm 15.7$)} \\
        \bottomrule
        \end{tabular}
\end{table}

\vspace{-10pt}


\subsection{Comparisons Analysis}

\begin{wrapfigure}{r}{0.5\textwidth} 
\vspace{-0.5cm}
  \centering
  \includegraphics[width=\linewidth]{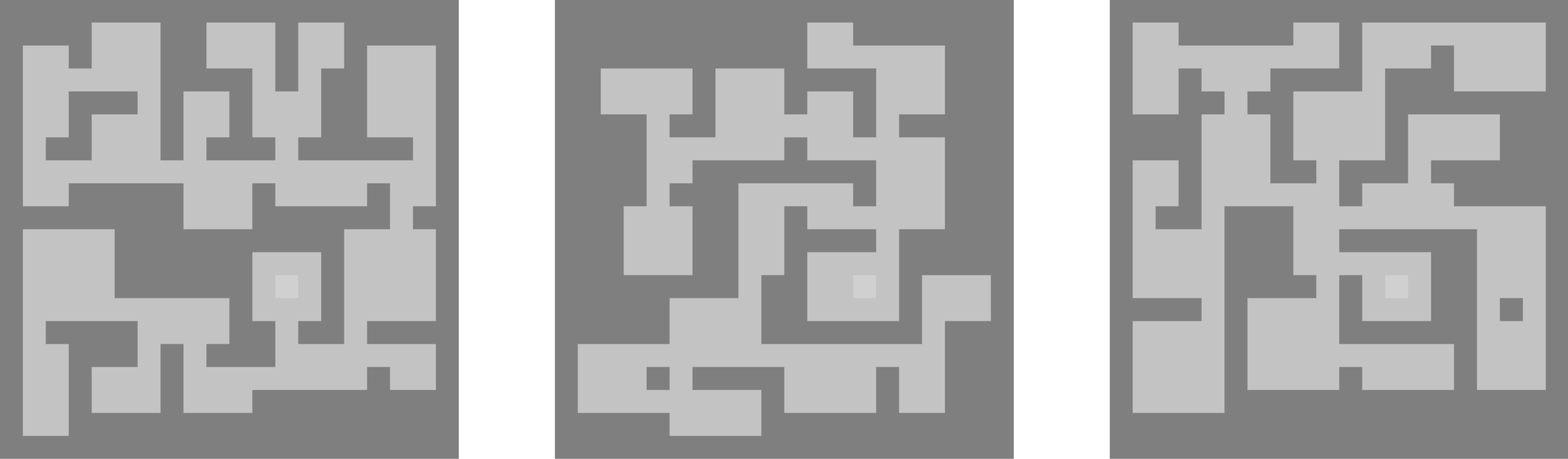}
  \caption{\textbf{examples maps from our testing set.}}
  \label{fig: test set}
\vspace{-0.5cm}
\end{wrapfigure}
As shown in Figure~\ref{fig: test set}, we construct 100 dungeon maps (100m $\times$ 100m) for benchmark testing which have never been seen by the agents during training.
We first investigate the performance of the agents' learned messages $\mu_i^t$ for information sharing.
As illustrated in Table~\ref{tab:exp_0.95}, when the team size $N = 4$, our model saves $\textbf{99.2\%}$ communication volume and surpasses \textbf{mtsp\_based} by $\textbf{2.4\%}$ in terms of travel distance.
These results indicate that learned messages can help robots reason about the global knowledge/state to generate high-quality, distributed exploration paths, without explicitly sharing detailed maps.
We further train a version of our DRL-based model, where robots are allowed to both share learned messages and explicit partial map. 
Even when sharing full belief maps, our variant \textbf{global\_map} still surpasses \textbf{mtsp\_based} by 9.2\% in terms of travel distance.
This demonstrates that our approach also exhibits exploration efficiency in high-bandwidth scenarios, where sharing maps among robots is possible.

Finally, we note that the scalability of our approach to larger team sizes is an important criterion for a distributed multi-robot system.
Without retraining, we double the team size $N$ from $4$ to $8$.
As demonstrated in Table~\ref{tab:exp_0.95}, our approach once again outperforms \textbf{mtsp\_based} by \textbf{$7.2\%$} in terms of travel distance.
Furthermore, the travel distance margin between \textbf{global\_map} and our approach still remains similar, even with this larger team size (only $11.1\%$ higher than \textbf{global\_map} when $N=8$).
We believe that these result demonstrate the robust scalability of our approach.


\section{Conclusion}

In this work, we propose a DRL-based multi-robot exploration framework that achieves a significant reduction in bandwidth consumption, while maintaining high exploration efficiency.
Our approach leverages communication learning to embed the most salient information from individual beliefs (partial map) of the environment into fixed-sized messages that are shared with other robots.
Robots then reason about their own belief as well as received messages to form an \textit{implicit}, global context of the environment (and overall state of the exploration task) and generate distributed exploration paths.
To these ends, our approach leverages privileged learning based on learned attention mechanisms, to endow the critic network with ground truth map knowledge and better guide the policy network during training.
Compared to our baselines, we demonstrate $99.2\%$ communication bandwidth savings while only sacrificing a marginal $2.4\%$ in travel distance.
These results highlight the capability of our robots to understand and reason about the current global state of the exploration task, without explicitly relying on other robots' detailed belief map. 

Future works will focus on further reducing the volume of communication within the team, to minimize bandwidth by reducing the frequency of communication autonomously or shrinking message sizes.
We will also further enhance the performance of our model with more efficient communication, and consider cases with intermittent communications.
Finally, we hope to conduct real-world experiments to validate the feasibility of our method in the future.

\section{Acknowledgements}
This work was supported by Temasek Laboratories (TL@NUS) under grant TL/FS/2022/01 and the Singapore Ministry of Education Academic Research Fund Tier 1.


\bibliographystyle{plain}
\bibliography{references}

\end{document}